\documentclass[conference]{IEEEtran}
\IEEEoverridecommandlockouts
\usepackage{cite}
\usepackage{amsmath,amssymb,amsfonts}
\usepackage{algorithmic}
\usepackage{graphicx}
\usepackage{textcomp}
\usepackage{xcolor}
\def\BibTeX{{\rm B\kern-.05em{\sc i\kern-.025em b}\kern-.08em
    T\kern-.1667em\lower.7ex\hbox{E}\kern-.125emX}}

\usepackage{caption}
\usepackage{amssymb}
\usepackage{mathtools}
\usepackage{hyperref}
\usepackage{todonotes}

\usepackage{booktabs}
\usepackage{multirow}
\usepackage{adjustbox}
\usepackage{colortbl}

\usepackage{listings}

\begin{document}

\title{Progressing from Anomaly Detection to Automated Log Labeling and Pioneering Root Cause Analysis}

\author{
Thorsten Wittkopp$\empty^1$, Alexander Acker$\empty^2$, and Odej Kao$\empty^1$\\
$\empty^1$Technische Universität Berlin, Germany, \{t.wittkopp, odej.kao\}@tu-berlin.de\\
$\empty^2$logsight.ai \{alexander.acker@logsight.ai\}
}

%\author{
%\IEEEauthorblockN{Thorsten Wittkopp, Alexander Acker and Odej Kao}
%\IEEEauthorblockA{Technische Universit{\"a}t Berlin, Germany, %\{firstname.lastname\}@tu-berlin.de}
%}

\maketitle

\begin{abstract}
The realm of AIOps is transforming IT landscapes with the power of AI and ML. Despite the challenge of limited labeled data, supervised models show promise, emphasizing the importance of leveraging labels for training, especially in deep learning contexts.
This study enhances the field by introducing a taxonomy for log anomalies and exploring automated data labeling to mitigate labeling challenges. It goes further by investigating the potential of diverse anomaly detection techniques and their alignment with specific anomaly types.
However, the exploration doesn't stop at anomaly detection. The study envisions a future where root cause analysis follows anomaly detection, unraveling the underlying triggers of anomalies. This uncharted territory holds immense potential for revolutionizing IT systems management.
In essence, this paper enriches our understanding of anomaly detection, and automated labeling, and sets the stage for transformative root cause analysis. Together, these advances promise more resilient IT systems, elevating operational efficiency and user satisfaction in an ever-evolving technological landscape.
\end{abstract}


\section{Introduction}
The complexity of modern IT systems presents operation and development teams with challenges in terms of both implementation and maintenance~\cite{rosendo2018improve}. 
To address this issue, the area of artificial intelligence for IT operations (AIOps) has emerged~\cite{gulenko2020ai}, which aims to support users in troubleshooting and problem mitigation.
AIOps systems rely on the three pillars of observability: metrics, traces, and logs~\cite{sridharan2018distributed}.

A key use case for AIOps is to support DevOps teams in detecting anomalies, resolving system failures, and mitigating their root causes. 
Due to the heavy use of logging in modern systems, many works have hereby focused on log anomaly detection~\cite{zawawy2010log,bogatinovski2020multi,wittkopp2021loglab,DBLP:conf/hicss/WittkoppSWAK23,hamooni2016logmine,korzeniowski2022landscape,lu2017log}. 
These can generally be divided into two classes: unsupervised~\cite{du2017deeplog,nedelkoski2020self,wittkopp2021a2log} and supervised~\cite{zhang2019robust,yang2021semi}.

One of the main obstacles in log anomaly detection is the lack of labeled log data~\cite{wen2020time} to train sufficient anomaly detection methods~\cite{yang2021semi}.
Labeling data is resource-intensive and time-consuming, requiring experts to examine each log message and identify associated errors. 
Supervised models trained on extensive datasets containing both normal and abnormal instances exhibit remarkable log anomaly detection performance~\cite{nedelkoski2020self,zhang2019robust}. 
Utilizing labels, regardless of their granularity, for training is advantageous. Labeled training data is particularly crucial for deep learning approaches and significantly enhances log anomaly detection acceleration~\cite{ratner2016data}.
Moreover, a critical catalyst for advancing anomaly detection techniques is the comprehension of the specific types of anomalies present within the log data.
A less explored area lies in the stages that follow anomaly detection, specifically in the realm of root cause analysis. Investigating this field has the potential to offer a valuable understanding of the underlying factors triggering these anomalies.

Hence, this study demonstrates the categorization of anomalies into distinct classes and the composition of known datasets corresponding to these anomalies. Moreover, it delves into how diverse anomaly detection techniques identify these distinct anomalies~\cite{wittkopp2021taxonomy}. To address the challenge of unlabeled data, an approach is presented for automated data labeling~\cite{wittkopp2021loglab}. Lastly, this paper explores the vision of identifying root causes for failures.

The remainder of this paper is structured as follows.
\autoref{sec:1} provides an overview of supervised, unsupervised, and weak-supervised methods.
\autoref{sec:2} presents PU learning to solve log analysis tasks.
\autoref{sec:3} points out our log anomaly taxonomy.
\autoref{sec:4} evaluates our automated log labeling approach LogLAB in comparison to nine other approaches.
\autoref{sec:5} envision a root cause analysis strategy with PU learning.
Lastly \autoref{sec:6} concludes the paper.

\section{Unsupervised, Weak-Supervised and Supervised Log Analysis}\label{sec:1}
Anomaly detection, a fundamental task in log analysis, aims to identify instances that deviate significantly from the norm or expected behavior in an IT system. Two primary approaches used for anomaly detection are supervised and unsupervised methods, each with its own strengths and limitations.

Supervised anomaly detection involves training a model on a labeled dataset that includes both normal and anomalous instances. The model learns to differentiate between the two classes based on the provided labels. During training, the model gains insights into the characteristics of normal instances and is then capable of identifying anomalies based on deviations from the learned normal behavior.

The advantages of supervised methods are:
\textbf{Clear Labeling:} Since supervised methods require labeled training data, the anomalies' true nature is known, leading to accurate detection.
\textbf{Effective for Known Anomalies:} Supervised methods are well-suited for scenarios where the types of anomalies are known and can be labeled.

Whereas the disadvantages are:
\textbf{Labeling Effort:} Labeled data is often expensive and time-consuming to obtain, particularly for rare or complex anomalies.
\textbf{Limited to Known Anomalies:} These methods struggle to detect novel or previously unseen anomalies, as they rely on patterns learned from existing data.

Common supervised methods are:
SVMs are well-recognized in anomaly detection~\cite{manevitz2001one,liang2007failure}, often ranking among the top performers. LogRobust~\cite{zhang2019robust} addresses stability concerns in existing methods using an attention-based Bi-LSTM model and semantic vectorization. SwissLog~\cite{LiCJHY20} advances this concept by incorporating semantic and temporal information for diverse fault handling. LogBERT~\cite{GuoYW21} expands BERT for log anomaly detection via tailored self-supervised training tasks. Similarly, Logsy~\cite{nedelkoski2020self} employs an attention-based encoder model, augmented with anomaly samples from auxiliary log datasets for improved vector representations. PLELog~\cite{YangCWWJDZ21} introduces an attention-based GRU neural network with PU learning and probabilistic label estimation.

Unsupervised anomaly detection, on the other hand, involves identifying anomalies without labeled data. The algorithm's objective is to uncover patterns that deviate from the majority of the data points. The idea is to identify instances that are significantly different from the typical behavior observed in the dataset.

Their advantages are:
\textbf{No Labeling Requirement:} Unsupervised methods are more flexible as they don't depend on labeled data, making them suitable for cases where labeling is impractical.
\textbf{Detecting Unknown Anomalies:} These methods can potentially uncover novel or emerging anomalies that were not encountered during training.

Whereas the disadvantages are:
\textbf{Lack of True Labels:} The absence of labeled data makes it challenging to validate the accuracy of detected anomalies.
\textbf{Subjectivity in Thresholds:} Determining the threshold for what constitutes an anomaly can be subjective and require domain expertise.

Common unsupervised methods are:
DeepLog~\cite{du2017deeplog} employs LSTM with templates~\cite{he2017drain} to treat logs as sentence sequences for anomaly detection. LogAnomaly~\cite{MengLZZPLCZTSZ19} integrates template2vec representations with LSTM networks to identify sequential and quantitative anomalies. In~\cite{FarzadG20}, Isolation Forests are combined with multiple Autoencoder Networks.

\subsection{Log Processing Techniques}
Irrespective of the model type, the majority of models utilize one of the subsequent techniques for log file analysis.
The log file records software executions via log instructions (e.g., \texttt{printf()} or \texttt{log.info()}). Each instruction generates a log message, forming a sequence $L = ( l_i : i=1,2,\ldots)$. Meta-information (timestamps, severity) and content (execution description) are typically distinguished. Common log processing techniques that are used on these logs are:

\textbf{Tokenization.} This process segments text (e.g., into words or characters). The smallest unit is a token, rendering log content as a token sequence $s_i = (w_j: w_j \in V, j = 1,2,3,\ldots)$. Here, $w$ signifies a token, $V$ is the token set (vocabulary), and $j$ denotes a token's position.

\textbf{Templates.} Log messages can become \emph{log templates} or \emph{log keys}, common in log anomaly detection~\cite{nedelkoski2020self,du2017deeplog}. Static tokens in a log form template $t_i$ for the i-th message. Unique templates have an id $x$ ($t^x$). The remaining tokens become attributes $a_i$ for the log message $l_i$. 
The log lines \textit{Start mail service at node wally001} and
\textit{Start printer service at node wally005} are processed to the template \texttt{'Start * service at node *'} with attribute sets \texttt{['mail', 'wally001']} and \texttt{['printer', 'wally005']} respectively. 

\textbf{Embedding}
An embedding $\vec{e_i}$ is a real-valued vector for a token, serving as a machine learning model input. Transformation function $g$ converts token sequence $c_i$ with length $s_i$ into a sequence of embeddings $\vec{e_i}$, with $g: V^{|s_i|} \rightarrow \mathbb{R}^{d,|s_i|}$. Embeddings adapt during training to represent original token semantics. Each $j$-th embedding in $\vec{e_i}$ is $\vec{e_i}(j)$, computed for each token $w_j$ in sequence $c_i$. Truncated embedding sequences $\vec{e}_i'$ are neural network inputs.

\section{PU Learning}\label{sec:2}
PU learning is a powerful approach that merges the strengths of supervised and unsupervised methods. It's particularly beneficial when accurate labeling is difficult or unnecessary. Thereby, PU learning offers a versatile solution for anomaly detection, data labeling, and root cause analysis, bypassing the need for extensive labeling and enhancing overall efficiency.
A critical step for PU learning is estimating time windows when important log entries occur. Accurate time windows enhance the effectiveness by providing focused context for log analysis.

We describe our log analysis tasks as weak supervision learning problems with inaccurate labels since we cannot assign accurate labels. Weak supervision with inaccurate labels is defined as a situation where the supervision information is not always matching the ground-truth~\cite{zhou2018brief}. 
Therefore, we assign preliminary abnormal labels for all log events in the estimated time windows and preliminary normal labels for all other log events.
We utilize PU learning~\cite{liu2002partially,liu2003building} in our log analysis methods, which is short for learning from positive and unlabeled data.
PU learning is an umbrella term for several weakly supervised binary classification methods that classify unlabeled samples by learning the positive (normal) and treating the unlabeled as abnormal \cite{liu2002partially,zhu2009introduction,bekker2020learning}.

\subsection{Model and Objective Function}
\begin{figure}[htbp]
\centering
\includegraphics[width=0.7\columnwidth]{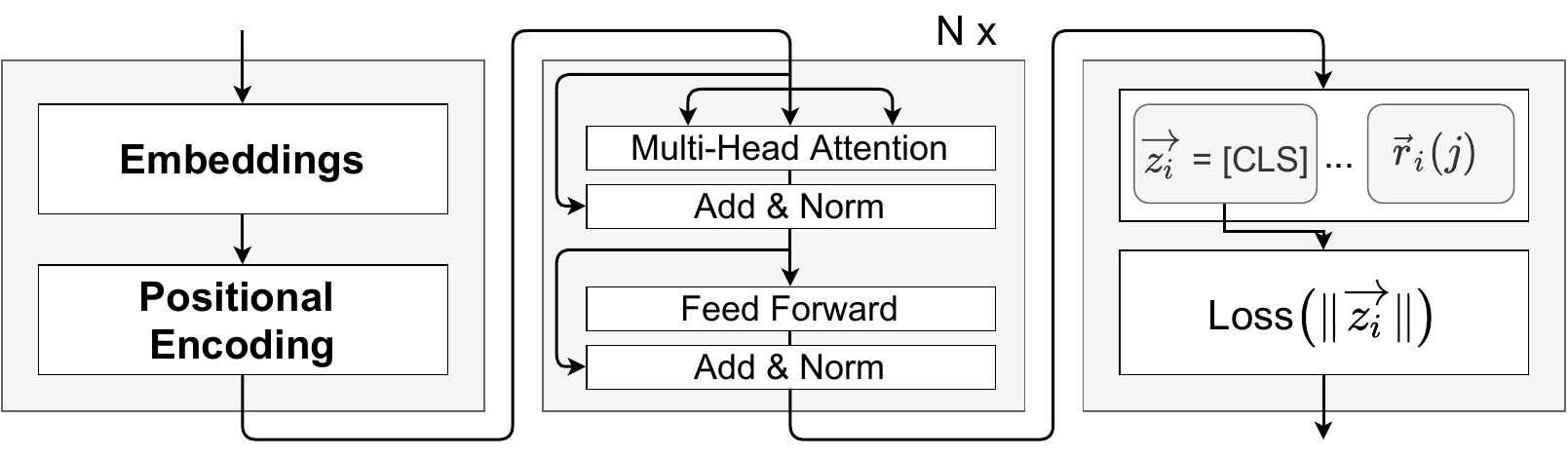}
\caption{Attention-based encoder architecture}
\label{fig:model_architecture}
\end{figure}

To determine the anomaly or root cause score for all log lines, we utilize an encoder architecture with self-attention, as depicted in Figure~\ref{fig:model_architecture}.

As the network is expected to output a score for each log line, which can be used for further downstream tasks.
This objective function must ensure that log lines in $\mathcal{U}$ receive scores if they are significantly different from log lines observed in $\mathcal{P}$.
On the other hand, the objective function should assign low scores to log lines that occur in both $\mathcal{P}$ and $\mathcal{U}$. 
Furthermore, the objective function should be able to handle a large number of mislabeled log lines that, by design, occur in the unknown class $\mathcal{U}$. 

To fulfill these requirements, we compute scores based on the Euclidean distance, representing the length of the output vector $\lVert z_i\rVert$ for each input sequence $\vec{e_i}'$. 
The objective function comprises two parts: the first part minimizes errors for samples in class $\mathcal{P}$ to yield small scores near zero, while the second part amplifies errors for samples in class $\mathcal{U}$ to drive higher scores. 
This structure is depicted in Equation \ref{eq:objective_function_gen}, where $\tilde{y}_i$ denotes the inaccurate label, $z_i$ represents the output vector of the model for each embedded input log message $\vec{e_i}'$, and $m$ indicates the number of samples per batch.

\begin{equation}
    \label{eq:objective_function_gen}
    \frac{1}{m}\sum\limits_{i=1}^{m}((1-\tilde{y}_i)*a(z_i) + (\tilde{y}_i)*b(z_i)
\end{equation}

For $a$, we minimize the error for positive samples and in contrast, we increase the error for all scores, when the log message is of class $\mathcal{U}$, with $a(z_i) = \lVert z_i \rVert^2$ and $b(z_i) = q^2 / \lVert z_i \rVert$, where $q$ is a numerator between 0 and 1 that represents the relation of the number of samples in $\mathcal{P}$ and $\mathcal{U}$.
Thus, the final objective function is composed as
\begin{equation}
    \frac{1}{m}\sum\limits_{i=1}^{n}\Big((1-y)*\lVert z_i \rVert^2 + (y)*\frac{(\frac{|\mathcal{P}|}{|\mathcal{P}|+|\mathcal{U}|})^2}{\lVert z_i \rVert}\Big)
\end{equation}

This effectively enables the encoder of the transformer model to train log messages with inaccurate labels by modifying the calculated error depending on the relation of $\mathcal{P}$ and $\mathcal{U}$.

\section{Error Types in Log Data}\label{sec:3}
Log data is vital for modern information systems, capturing diverse information about processes, events, and activities. Yet, anomalies can emerge within this data, deviating from the norm. Differentiating anomaly types is crucial for effective anomaly detection methods. Generally, each data sample is defined by its features, which vary by domain. For instance, time series data relies on position and time, while natural language processing uses word vectors. Established anomaly taxonomy categorizes anomalies into \emph{Point Anomalies} and \emph{Contextual Anomalies}~\cite{chandola2009anomaly,sebestyen2018taxonomy}.

\begin{figure}[htbp]
\centering
\includegraphics[width=0.7\columnwidth]{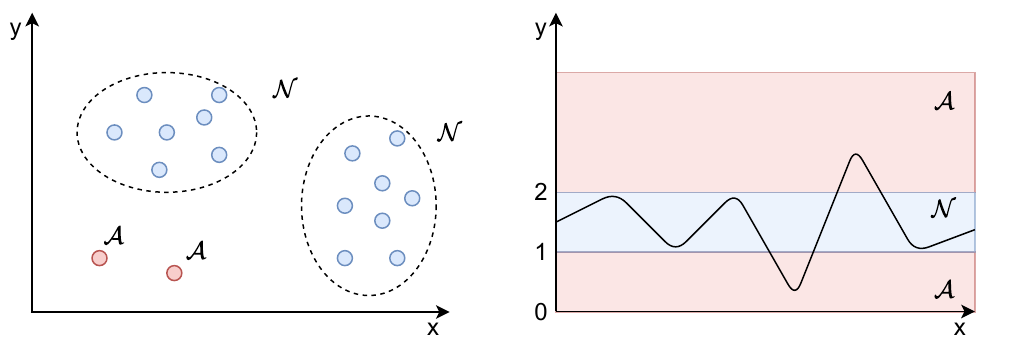}
\caption{Two examples for point anomalies. On the left side: Point anomalies in 2D space. On the right side: Point anomalies in a time series.}
\label{fig:point_anomalies_examples}
\end{figure}

\textbf{Point Anomalies.} 
A point anomaly is a lone data sample that stands out from the rest~\cite{chandola2009anomaly}, with its feature set values markedly diverging from those of normal samples $\mathcal{N}$.
Examples of point anomalies are shown in \autoref{fig:point_anomalies_examples}. In the first instance, anomalies lie beyond the normal sample zone in a 2D space. In the second, time series data has anomalies that fall outside the usual $y\in[1,2]$ range for the feature set.

\begin{figure}[htbp]
\centering
\includegraphics[width=0.45\columnwidth]{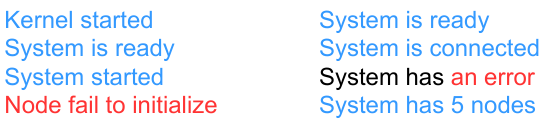}
\caption{Two examples for point anomalies in written text.}
\label{fig:point_anomalies_examples_nlp}
\end{figure}

In \autoref{fig:point_anomalies_examples_nlp}, two text-based point anomalies are displayed. In the first case, the anomaly, "Node failed to initialize," stands apart from other sentences in terms of its word features. The second instance is more nuanced, with specific words marking the anomaly. While sentences share a common or similar prefix, only the later descriptors ("ready, connected, 5 nodes, an error") pinpoint the anomaly.

\textbf{Contextual Anomalies.} Contextual anomalies~\cite{chandola2009anomaly}, also referred to as conditional anomalies~\cite{song2007conditional}, are samples anomalous only in a particular context. These anomalies might share the same feature set as normal samples but stand out due to their surrounding samples.

\begin{figure}[htbp]
\centering
\includegraphics[width=0.8\columnwidth]{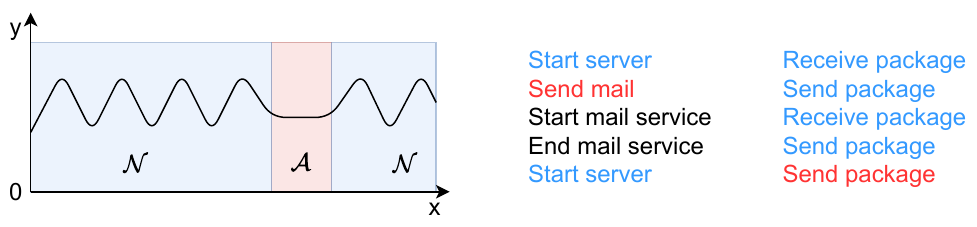}
\caption{One example of a contextual anomaly in time series and two examples for contextual anomalies in written text.}
\label{fig:context_anomalies_examples_both}
\end{figure}

In \autoref{fig:context_anomalies_examples_both}, anomalies with normal y-values disrupt the context, altering established patterns. This is highlighted in two text examples. In the left case, the context of \textit{Send mail} preceded by \textit{Start mail service} and followed by \textit{End mail service} makes it a contextual anomaly. Similarly, in the second example, alternating \textit{Receive package} and \textit{Send package} is interrupted by an extra \textit{Send package} statement, forming another contextual anomaly.

\subsection{Classifying Anomalies in Log Data}
Our classification involves \textit{Point Anomalies} and \textit{Contextual Anomalies}. Additionally, we differentiate between \textit{Template Anomalies} and \textit{Attribute Anomalies} within point anomalies.

\begin{figure}[htbp]
\centering
\includegraphics[width=0.5\columnwidth]{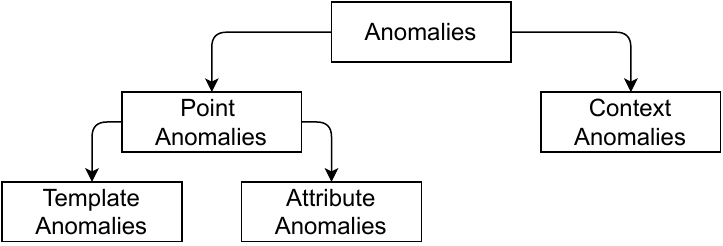}
\caption{Taxonomy for anomalies in log data.}
\label{fig:anomaly_taxonomy}
\end{figure}

\autoref{fig:anomaly_taxonomy} depicts our taxonomy. 
In log data, a \textit{Point Anomaly} is an anomaly within a log message itself. It's identified independently, without considering context. This could be indicated by a template or specific word/number (an attribute) in the message. We term a \textit{Template Anomaly} based on the message's template, while an \textit{Attribute Anomaly} relates to attributes generated from the template.

The second type is \textit{Contextual Anomalies}. These depend on neighboring log messages to define anomaly. An individual message's content is significant within its surrounding context.

\subsection{Anomaly Classification Method}

\begin{figure}[htbp]
\centering
\includegraphics[width=1.0\columnwidth]{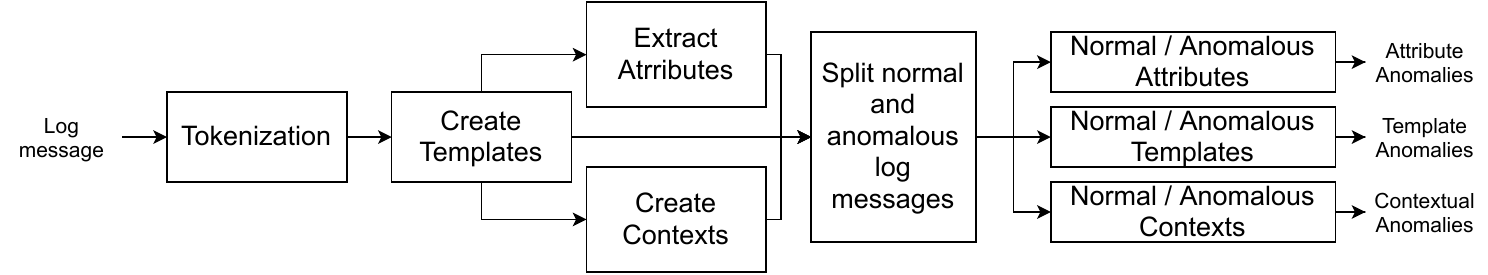}
\caption{Mining process of the different anomaly types.}
\label{fig:classification_process}
\end{figure}

Illustrated in \autoref{fig:classification_process}, our taxonomy's anomaly classification process begins by segmenting each log message into token sequences, and mining log templates. Attributes are then extracted once all templates are generated. Subsequently, context is computed for each log line. Context $c_i$ for log message $l_i$ relies on template ids and forms a set $c_i = { t^x_j : j = i-a,\ldots,i-1,i+1,\ldots,i+b] }$, where $a$ and $b$ set context boundaries. For instance, with boundaries $a=2$ and $b=1$, the context of the 10th log message is $c_{10}={l_8,l_9, l_{11}}$, excluding its own template.

Post template, attribute, and context derivation, the dataset splits into normal log messages $\mathcal{N}$ and anomalous log messages $\mathcal{A}$. Using these, we assign a score to each message for each anomaly type, gauging its anomaly strength. Scores fall within $[0,1]$, where $1$ signifies the highest manifestation.

\textbf{Template Anomalies.}
For each template id $x$, we compute the template anomaly $\alpha$. Denoted by $t^x(\cdot)$, all templates with id $x$ are collected. The formula is:
\begin{equation}
\alpha(t^x) = \frac{|t^x(\mathcal{A})|}{|t^x(\mathcal{A})|+|t^x(\mathcal{N})|}
\label{eq:template_anomaly}
\end{equation}

\textbf{Attribute Anomalies.}
For each log message $l_i$, we calculate the attribute anomaly $\beta$. Given that a message may have multiple attributes, we find the highest score by evaluating each attribute's contribution:
\begin{equation}
\beta(a_i) = \max{(s: \forall a_j \in a_i. s= \frac{|a_j(\mathcal{A})|}{|a_j(\mathcal{A})|+|a_j(\mathcal{N})|})}
\label{eq:attribute_anomaly}
\end{equation}

\textbf{Contextual Anomalies.}
Per log message $l_i$, the contextual anomaly $\gamma$ is computed. Represented as $c_i(\cdot)$, the same contexts for context $c_i$ are grouped:
\begin{equation}
\gamma(c_i) = \frac{|c_i(\mathcal{A})|}{|c_i(\mathcal{A})| + |c_i(\mathcal{N})|}
\label{eq:context_anomaly}
\end{equation}
Thus, scores are determined by dividing anomalous event occurrences by the cumulative occurrences in both sets.

\subsection{Analysis of Benchmark Datasets}
We analyze Thunderbird, Spirit, and BGL datasets to grasp anomaly distribution in our taxonomy. Additionally, five key unsupervised log anomaly detection methods are trained on these datasets for diverse anomaly prediction evaluation.

These evaluation datasets are from various large-scale computer systems, expert-labeled as described in \cite{oliner2007datasets}. Details about normal and anomalous log counts, template counts, and intersecting templates are in \autoref{table:datasets}. \emph{Thunderbird} has 211+ million logs from a Sandia National Labs (SNL) supercomputer. \emph{Spirit}, from an SNL Spirit supercomputer, has 272+ million logs. The \emph{BGL} dataset, from a Lawrence Livermore National Labs (LLNL) BlueGene/L supercomputer, has 4,747,963 logs. Initial 5 million logs were taken from \emph{Thunderbird} and \emph{Spirit}.

\begin{table}[t]
	\centering
	\caption{Dataset Statistics. Templates were generated using Drain3~\cite{he2017drain}.}
	\begin{tabular}{p{1.3cm}cc p{0.3cm}c p{0.2cm} p{0.2cm} p{0.7cm}}
		\toprule
        \multirow{2}{*}{Dataset}  & \multicolumn{2}{c}{Log messages} & & \multicolumn{2}{c}{Templates} && \\
        \cmidrule{2-3} \cmidrule{5-8}
                &  normal & anorom. && normal & anorm. &&  intersec. \\
        \midrule
        Thunderbird   & 4\,773\,713 & 226\,287  && 969 & 17 && 3 \\
        Spirit        & 4\,235\,109 & 764\,891  && 1121 & 23 && 5 \\
        BGL           & 4\,399\,503 & 348\,460  && 802 & 58 && 10 \\
        \bottomrule
	\end{tabular}
	\label{table:datasets}
\end{table}

\begin{figure}[h]
\centering
\includegraphics[width=1\columnwidth]{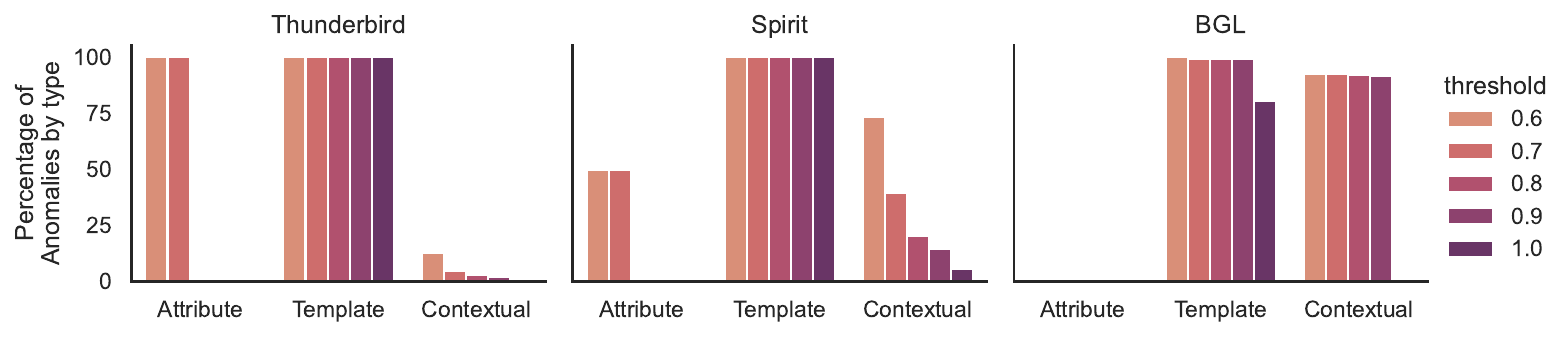}
\caption{Percentage of anomalies by type at different thresholds.}
\label{fig:approach_results}
\end{figure}

Our anomaly classification approach was applied to the datasets using thresholds of 0.6, 0.7, 0.8, 0.9, and 1.0, with context boundaries set at $a=10$ and $b=0$. Results are displayed in \autoref{fig:approach_results}. Even at threshold 1.0, over 99\% of anomalies in Thunderbird and Spirit datasets are classified as template anomalies, due to a small intersection between normal and abnormal templates. In Thunderbird, 99.9\% of anomalies share the same template, similarly 99.4\% in Spirit.

The scenario is similar for BGL, though its log templates are more varied. Only at threshold 1 do template anomalies drop to around 80\%.
Until threshold 0.7, nearly all Thunderbird anomalies are attribute anomalies. Higher thresholds lead to zero attribute anomalies, as some "anomalous" attributes are also present in normal log messages. For Spirit, one of two key log templates in Listing 1 contains an attribute, explaining the ~50\% attribute anomaly count, which drops to zero at higher thresholds. BGL has no attribute anomalies, but over 91\% of anomalies are contextual anomalies for thresholds 0.6 to 0.9, surpassing Thunderbird (13-2\%) and Spirit (63-14\%).

In summary, algorithms focusing on template anomalies should perform well in all three benchmark datasets. Detecting attribute anomalies has limited benefits due to their overlap with template anomalies. Context-based anomaly detection shows promise in BGL but is expected to be less effective for Thunderbird and moderately so for Spirit.

\subsection{Evaluation of Unsupervised Learning Methods}
We trained five unsupervised anomaly detection algorithms with a 0.7 threshold across all three datasets. This aimed to assess the ease of predicting different anomalies and the performance of each method. We chose Deeplog~\cite{du2017deeplog}, A2Log~\cite{wittkopp2021a2log}, PCA~\cite{he2016experience}, Invariant Miners~\cite{lou2010mining}, and Isolation Forest~\cite{liu2008isolation}. Evaluation covered four train/test splits: 0.2/0.8, 0.4/0.6, 0.6/0.4, and 0.2/0.8, testing method robustness.

\begin{figure}[bht]
\centering
\includegraphics[width=1\columnwidth]{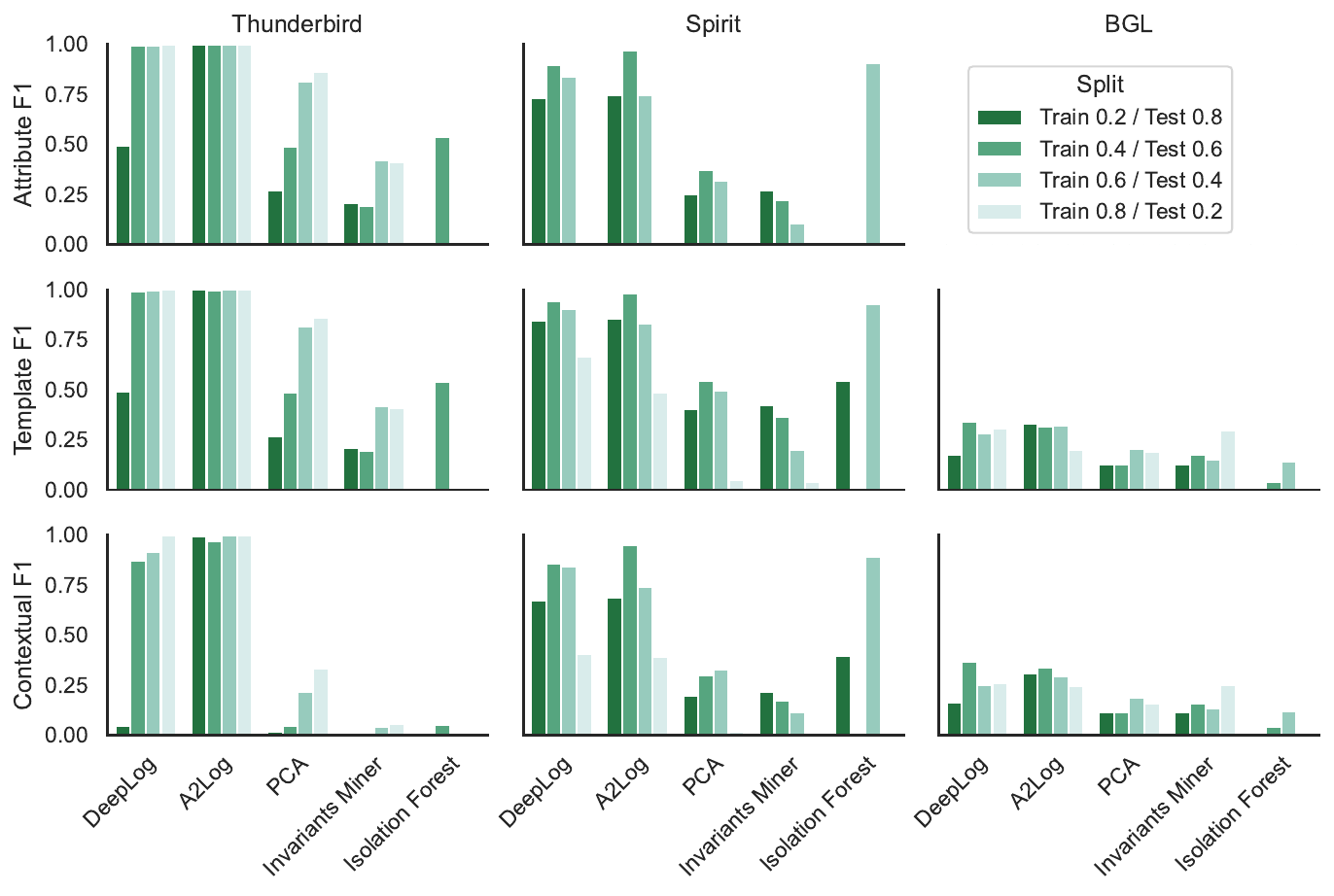}
\caption{F1 scores for predicting attribute, template, and contextual anomalies at different train/test splits at threshold 0.7.
BGL contains no attribute anomalies.}
\label{fig:ml_results}
\end{figure}

Results are depicted in \autoref{fig:ml_results}. Deep learning outperforms data mining in all experiments. Isolation Forest's sensitivity to specific splits excludes it from further analysis.
For Thunderbird, DeepLog and A2Log accurately classify most anomalies, with A2Log performing better. Non-deep learning struggles with attribute and template anomalies, as well as contextual anomalies.
On Spirit, deep learning achieves F1 scores around 0.75, 0.85, and 0.7 for attribute, template, and contextual anomalies.
F1 scores on BGL for template and contextual anomalies are low: DeepLog around 0.27, A2Log around 0.3, PCA around 0.14, and Invariants Miner around 0.15. Attribute anomalies are absent.
Unsupervised methods find template anomalies easiest to predict. Attribute anomalies without templates appear challenging. Deep learning reliably detects contextual anomalies, generally tougher than template anomalies.

\subsection{Summary}
The taxonomy will enable researchers and IT Operators to better understand their datasets and help them to pick suitable anomaly detection algorithms.
Future work should investigate the log messages our approach fails to classify, potentially hinting towards further classes that are currently not present in the taxonomy.

\section{Automated Log Labeling}\label{sec:4}
Automated log labeling is a significant task due to the time-intensive nature of manual labeling. This process proves especially valuable as it facilitates the development of supervised anomaly detection models. Moreover, it aids in the evaluation of various methods by providing labeled data for training and validation, thereby enhancing the accuracy and efficiency of log analysis techniques.

In order to determine time windows for applying PU learning and automating data labeling, we make use of monitoring solutions that generate alerts when anomalies, system disruptions, or failures arise. These alerts are triggered by monitoring various aspects such as metrics, hardware component issues, deployment failures, and other unusual scenarios~\cite{sukhwani2017monitoring}.

\begin{figure}[htbp]
\centering
\includegraphics[width=1.0\columnwidth]{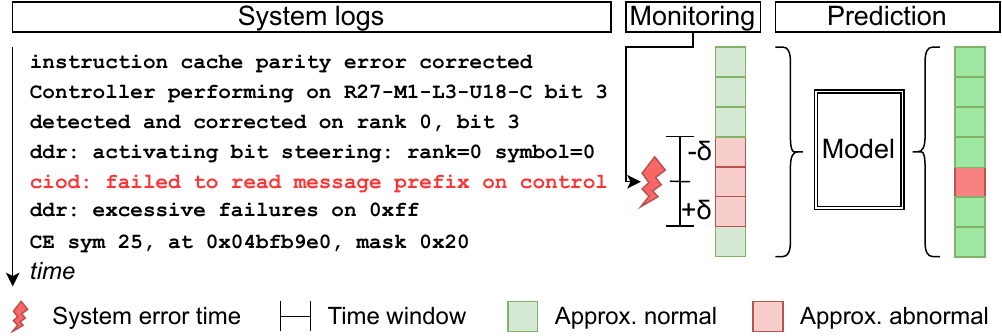}
\caption{We use rough estimates for failure times provided by monitoring systems in order to identify and label abnormal log messages via weak supervision.}
\label{fig_lab:problem_description}
\end{figure}

\autoref{fig_lab:problem_description} provides an example of the described problem.
It displays the log of a system with one abnormal log event (colored in red).
We utilize monitoring information to estimate time windows of the length $2*\delta$ in which we suspect abnormal log events. 
The model's task is to identify the abnormal log messages in the time window and classify all others as normal.

%We describe log labeling as a weak supervision learning problem with inaccurate labels.
Thereby, the underlying log data is divided into two classes, positive $\mathcal{P}$ and unlabeled $\mathcal{U}$, where $\mathcal{U}$ consists of all log messages that occur in the aforementioned time windows and $\mathcal{P}$ of the remaining log messages.

\subsection{LogLAB}
For the labeling of logs, we design a processing and modeling pipeline illustrated in~\autoref{fig:high_level_pipieline}. The individual steps are as follows:

\begin{figure}[htbp]
\centering
\includegraphics[width=1.0\columnwidth]{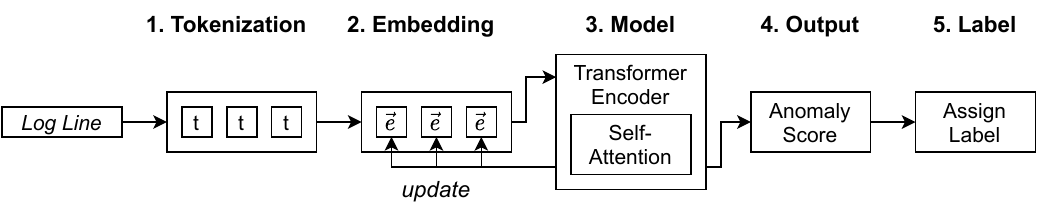}
\caption{High-level log message labeling pipeline.}
\label{fig:high_level_pipieline}
\end{figure}

First, we convert the content $c_i$ of each log message $l_i$ into a sequence of tokens $t_i$ by splitting on the symbols \texttt{.,:/} and whitespaces. 
Subsequently, we clean the resulting sequence of tokens by replacing certain tokens with placeholders. 
Thereby placeholder tokens for hexadecimal values \texttt{'[HEX]'} and any number greater or equal 10  \texttt{'[NUM]'} are introduced. Finally, we prefix the sequence of transformed tokens with a special token \texttt{'[CLS]'} which serves as a numerical summary of the whole log message. 

Since these sequences can vary in length, we truncate them to a fixed size and pad smaller sequences with \texttt{'[PAD]'} tokens.
For each token $w_j$ of the token sequence $t_i$, an embedding $\vec{e}_{i}(j)$ is obtained. 
The truncated sequences of embeddings $\vec{e'_i}$ serves as the input for the model.

The model computes an output embedding, for each input sequence $\vec{e'_i}$, which summarizes the log message by utilizing the embeddings of all tokens. 
This output embedding is encoded in the embedding of the \texttt{’[CLS]’} token which is also modified during training. 
For this purpose, we utilize the transformer architecture~\cite{DevlinCLT19} with additional self-attention~\cite{VaswaniSPUJGKP17}.
During the training process, the model is supposed to learn the meanings of the log messages, thereby getting an intuition of what is normal and abnormal.
Finally, this model outputs a vector (embedding) for each input sequence $\vec{e'_i}$. 
We denote the output of the model as $z_i=\Phi(\vec{e'_i};\Theta)$ and use it throughout the remaining steps.
Thereby the anomaly score is calculated by the length of the output vector $\lVert z_i \rVert$.
Anomaly scores close to $0$ represent normal log messages, whereby large vectors indicate an abnormal log message. 
The computed anomaly score is used to assign a label $\widehat{y_i}$ to the log message $l_i$, i.e. either normal or abnormal.

\subsection{Evaluation}\label{sec_lab:evaluation}
To obtain a significant and wide benchmark, we compare LogLAB to several state-of-the-art text-classification and anomaly detection approaches presented in a recent text-classification survey~\cite{kowsari2019text} as well as in an established survey for anomaly detection in system logs~\cite{he2016experience}.
Namely, we choose PCA, Invariant Miners, Deeplog, Decision Trees, Random Forests, SVMs, Logistic Regression, the Rocchio algorithm, and boosting approaches as benchmark methods. 
Thereby we measure the deviation from the ground truth $y_i$ and the calculated labels $\widehat{y_i}$.

\subsubsection{Experimental Setup}
We evaluate all methods on three labeled log datasets recorded at different large-scale computer systems\cite{oliner2007datasets}.
The \emph{BGL} dataset contains 4\,747\,963 log messages of which 7.3\,\% are abnormal and records a period of ~214 days, with on average 0.25 log messages per second.
We selected the first 5 M log messages from the \emph{Thunderbird} dataset of which 4.5\,\% are abnormal. They account for a period of ~9 days, with on average 6.4 log messages per second.
Again, we selected the first 5 M log messages from the \emph{Spirit} dataset of which 15.3\,\% are abnormal. They cover a period of ~48 days, with on average 1.2 log messages per second.

We create our evaluation datasets with inaccurate labels by including all abnormal log events as well as their surrounding events within a time window $2*\delta$ in $\mathcal{U}$; all remaining log events are in $\mathcal{P}$. 
Thereby we investigate the performance at three different time windows: $\pm 1000\,ms$ (2s), $\pm 5000\,ms$ (10s) and $\pm 15000\,ms$ (30s).
The amount of samples in $\mathcal{U}$ is changing for BGL: $0.39M$, $~0.44M$ and $~0.47M$,
Thunderbird: $1.42M$, $2.36M$ and $2.90M$ and 
Spirit: $1.00M$, $2.33M$ and $3.26M$ regarding the respective time window $\delta$.

Each sequence of tokens $t_i$ is truncated to have a length of 20 for \emph{Thunderbird}, 16 for \emph{Spirit}, and 12 for \emph{BGL}. 
The dimensionality $d$ of our embeddings is set to 128.
For the training of our LogLAB model, we use a hidden dimensionality of 256, a batch size of 1024, a total of 8 epochs, and a dropout rate of 10\%. 
We use the Adam optimizer with a learning rate of $10^{-4}$ and a weight decay of $5 \cdot 10^{-5}$.

\subsubsection{Results}

\begin{adjustbox}{
center,label={table:results},caption={Evaluation results: F1-scores above 0.99 and 0.98 are highlighted in blue and cyan, respectively.},float=table}
\scriptsize

\resizebox{0.5\textwidth}{!}{%
\begin{tabular}[h!]{p{0.5cm} p{0.8cm} p{0.5cm} p{0.5cm} p{0.5cm} p{0.5cm} p{0.5cm} p{0.5cm} p{0.5cm} p{0.5cm} p{0.5cm} p{0.8cm} } 
	\toprule
 	& & \multicolumn{3}{c}{Learning $\mathcal{P}$} & \multicolumn{7}{c}{Learning $\mathcal{P}$ and $\mathcal{U}$} \\
 	\cmidrule(lr){3-5}
 	\cmidrule(lr){6-12}
	\rotatebox{90}{Dataset} & Metric & \rotatebox{45}{PCA} & \rotatebox{45}{Invariant Miners} & \rotatebox{45}{Deeplog} & \rotatebox{45}{Decision Tree} & \rotatebox{45}{Random Forest} & \rotatebox{45}{SVM} & \rotatebox{45}{Logistic Regr.} & \rotatebox{45}{Boost} & \rotatebox{45}{Rocchio} & \rotatebox{45}{\textbf{LogLAB}} \\ [0.5ex]
	
 	\midrule
	\multicolumn{12}{c}{$\delta=\pm 1000ms$} \\
	\midrule
 
 	%\multirow{1}{*}{\rotatebox{0}{BGL}} 
 	%& Recall     & 1.0000 & 0.9999 & 0.6474 & 1.0000 & 1.0000 & 1.0000 & 0.9999 & 0.9898 & 0.6206 & 0.9954 \\
 	%& Precision  & 0.4248 & 0.3425 & 0.9680 & 0.9948 & 0.9666 & 0.9685 & 0.9952 & 0.9918 & 0.8286 & 0.9999 \\
 	BGL & F1-S.   & 0.5963 & 0.5102 & 0.7759 & \cellcolor{blue!20}0.9974 & \cellcolor{cyan!20}0.9830 & \cellcolor{cyan!20}0.9840 & \cellcolor{blue!20}0.9976 & \cellcolor{blue!20}0.9908 & 0.7096 & \cellcolor{blue!20}0.9977 \\ \midrule
 
	%\multirow{1}{*}{\rotatebox{0}{\parbox{1.1cm}{\centering\linespread{0.8}\selectfont thunder bird}}} 
	%& Recall     & 1.0000 & 1.0000 & 0.9224 & 1.0000 & 1.0000 & 0.9998 & 0.9998 & 0.9990 & 0.9990 & 0.9990 \\
 	%& Precision  & 0.1798 & 0.1003 & 0.0462 & 0.1934 & 0.1865 & 0.1930 & 0.1934 & 0.2020 & 0.2077 & 1.0000 \\
 	TBird & F1-S.   & 0.3048 & 0.1824 & 0.0880 & 0.3242 & 0.3144 & 0.3235 & 0.3242 & {0.3361} & 0.3440 & \cellcolor{blue!20}0.9995 \\ \midrule
 
 	%\multirow{1}{*}{\rotatebox{0}{spirit}} 
 	%& Recall     & 1.0000 & 1.0000 & 0.9999 & 1.0000 & 1.0000 & 0.9999 & 0.9994 & 0.9990 & 0.9943 & 0.9995 \\
 	%& Precision  & 0.6726 & 0.4092 & 0.9855 & 0.9935 & 0.9238 & 0.9719 & 0.9930 & 0.9945 & 1.0000 & 0.9999 \\ 
 	Spirit & F1-S.   & 0.8043 & 0.5807 & \cellcolor{blue!20}0.9926 & \cellcolor{blue!20}0.9967 & 0.9604 & \cellcolor{cyan!20}0.9857 & \cellcolor{blue!20}0.9962 & \cellcolor{blue!20}0.9968 & \cellcolor{blue!20}0.9971 & \cellcolor{blue!20}0.9997 \\ 
 	
 	\midrule
	\multicolumn{12}{c}{$\delta=\pm 5000ms$} \\
	\midrule
 	%& \multicolumn{3}{c}{Only learning P} & \multicolumn{7}{c}{P and U learning} \\ \midrule
 
 	%\multirow{3}{*}{\rotatebox{90}{BGL}} 
 	%& Recall     & 1.0000 & 0.9999 & 0.6371 & 0.9999 & 0.9999 & 0.9999 & 0.9999 & 0.9894 & 0.7614 & 0.9915 \\
 	%& Precision  & 0.4214 & 0.3434 & 0.9908 & 0.9753 & 0.9317 & 0.9381 & 0.9755 & 0.9698 & 0.8549 & 0.9984 \\
 	BGL & F1-S.   & 0.5930 & 0.5112 & 0.7755 & \cellcolor{cyan!20}0.9874 & 0.9646 & 0.9680 & \cellcolor{cyan!20}0.9875 & 0.9795 & 0.8054 & \cellcolor{blue!20}0.9949 \\ \midrule
 
	%\multirow{3}{*}{\rotatebox{90}{\parbox{1.1cm}{\centering\linespread{0.8}\selectfont thunder bird}}} 
	%& Recall     & 1.0000 & 1.0000 & 0.2070 & 0.9999 & 1.0000 & 0.9999 & 0.9998 & 0.9991 & 0.9998 & 0.9990 \\
 	%& Precision  & 0.1801 & 0.1072 & 0.0386 & 0.1540 & 0.1373 & 0.1389 & 0.1546 & 0.1675 & 0.1866 & 1.0000 \\
 	TBird & F1-S.   & 0.3053 & 0.1936 & 0.0651 & 0.2669 & 0.2415 & 0.2439 & 0.2678 & 0.2869 & 0.3146 & \cellcolor{blue!20}0.9995 \\ \midrule
 
 	%\multirow{3}{*}{\rotatebox{90}{spirit}} 
 	%& Recall     & 1.0000 & 1.0000 & 0.9998 & 0.9999 & 1.0000 & 1.0000 & 0.9999 & 1.0000 & 0.9943 & 0.9959 \\
 	%& Precision  & 0.6249 & 0.4025 & 0.9862 & 0.4829 & 0.3749 & 0.3874 & 0.4829 & 0.4115 & 0.9949 & 0.9999 \\
 	Spirit & F1-S.   & 0.7691 & 0.5740 & \cellcolor{blue!20}0.9929 & 0.6513 & 0.5453 & 0.5584 & 0.6560 & 0.5830 & \cellcolor{blue!20}0.9946 & \cellcolor{blue!20}0.9980 \\
 	
 	\midrule
	\multicolumn{12}{c}{$\delta=\pm 15000ms$} \\
	\midrule
 	%& \multicolumn{3}{c}{Only learning P} & \multicolumn{7}{c}{P and U learning} \\ \midrule
 	%\multirow{3}{*}{\rotatebox{90}{BGL}} 
 	%& Recall     & 1.0000 & 0.9999 & 0.6384 & 0.9999 & 0.9999 & 0.9999 & 0.9999 & 0.9851 & 0.7631 & 0.9922 \\
 	%& Precision  & 0.4164 & 0.3450 & 0.9893 & 0.9518 & 0.9018 & 0.9090 & 0.9545 & 0.9674 & 0.8185 & 0.9883 \\
 	BGL & F1-S.   & 0.5879 & 0.5130 & 0.7760  & 0.9753 & 0.9483 & 0.9523 & 0.9767 & 0.9762 & 0.7898 & \cellcolor{blue!20}0.9902 \\ \midrule
 
	%\multirow{3}{*}{\rotatebox{90}{\parbox{1.1cm}{\centering\linespread{0.8}\selectfont thunder bird}}} 
	%& Recall     & 1.0000 & 1.0000 & 0.3681 & 0.9999 & 1.0000 & 0.9999 & 0.9999 & 0.9991 & 0.9998 & 0.9990 \\
 	%& Precision  & 0.1782 & 0.1070 & 0.0656 & 0.0719 & 0.0666 & 0.0723 & 0.0723 & 0.0797 & 0.1262 & 1.0000 \\
 	TBird & F1-S.   & 0.3025 & 0.1933 & 0.1113 & 0.1341 & 0.1248 & 0.1348 & 0.1350 & 0.1476 & 0.2241 & \cellcolor{blue!20}0.9995 \\ \midrule
 
 	%\multirow{3}{*}{\rotatebox{90}{spirit}} 
 	%& Recall     & 1.0000 & 1.0000 & 0.9999 & 0.9999 & 1.0000 & 1.0000 & 1.0000 & 0.9999 & 0.9944 & 0.9579 \\
 	%& Precision  & 0.4256 & 0.3984 & 0.7001 & 0.2220 & 0.2102 & 0.2222 & 0.2221 & 0.2063 & 0.3740 & 0.9240 \\ 
 	Spirit & F1-S.   & 0.4958 & 0.4254 & 0.8236 & 0.4909 & 0.4735 & 0.4836 & 0.4917 & 0.4887 & 0.5192 & \cellcolor{cyan!20}0.9825 \\ 
 	\bottomrule
\end{tabular}
}
\end{adjustbox}

To compare LogLAB to our baselines, we assess the prediction performance $\tilde{y}_i \sim y_i$ in terms of F1-score metrics. The F1-scores are presented in \autoref{table:results}. 
As expected, with increasing $\delta$ and thus growing size of $\mathcal{U}$, the performance across all approaches tends to decrease.

For $\delta = \pm 1000\,ms$ this was apparently easy to achieve for most of the methods. An exception is the Thunderbird dataset, which is characterized by a large $\mathcal{U}$ class: No baseline manages to achieve an F1-score higher than 0.35, except LogLAB.

For $\delta = \pm 5000\,ms$ we notice that the performance degradation previously observed for most approaches on the Thunderbird dataset now also start to manifest on the Spirit dataset. 
The biggest gap in performance becomes evident at the largest observed time window of $\delta = \pm 15000\,ms$. 
For the dataset BGL, we notice a considerable drop in F1-scores of other approaches to 0.97, while LogLAB maintains its high performance.

\subsection{Summary}
Introducing LogLAB: an innovative model automating log data labeling, sidestepping manual expert efforts. Using estimated time windows from monitoring systems, it generates labeled datasets. Employing attention mechanisms and a tailored weak supervision objective, LogLAB handles data imbalance and imprecise labels. Our evaluation against nine benchmarks on three datasets showcases LogLAB's dominance, yielding over 0.98 F1-score, even with numerous inaccurate labels. Future enhancements may involve iterative label adjustment and integrating alternative time window estimation sources for improved accuracy..

\section{Root Cause Analysis}\label{sec:5}
In this chapter, we want to envision how a root cause analysis could be performed with our PU learning approach and what difficulties there are in doing so.
Root cause analysis serves as a subsequent phase following the initial anomaly detection process. Its primary goal is to pinpoint the underlying triggers that led to a specific system failure.
Conversely, log-based root cause analysis strives to choose a minimal, contextually related set of log lines. This aids development or operations teams in comprehending true failure causes, a task involving insight into the actual course of action~\cite{zawawy2010log}. 

Root cause analysis faces three key challenges due to a lack of root cause information:

\begin{enumerate}
    \item Identifying root cause log lines within a time window is uncertain. This leads to a high number of mislabeled training data as all log lines before a failure are potential root cause candidates.
    \item The number of log lines indicating a root cause varies, unlike the binary classification in traditional anomaly detection. Our aim is to identify an unknown number of log lines.
    \item Training data is often imbalanced, with some root causes being common while others are rare or absent. This can create training biases, affecting model performance.
\end{enumerate}

\subsection{Root Cause Analysis as a PU Learning Problem}
\label{sec:PU_learning}
\begin{figure}[htbp]
\centering
\includegraphics[width=1\columnwidth]{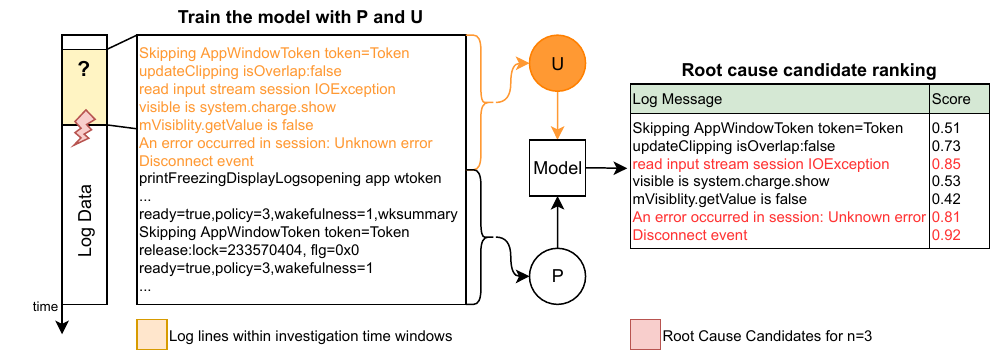}
\caption{Illustrating training with false labels and desired outcome for n=3. Orange lines are $\mathcal{U}$ class, and black lines are $\mathcal{P}$.}
\label{fig:problem_example}
\end{figure}

First, root cause analysis approaches demand failure detection. These can be identified via anomaly detection methods or other monitoring systems. Next, system experts should select a suitable \emph{time window} before the failure, where root cause log lines are anticipated since they know the system best. In Figure~\ref{fig:problem_example}, such a window (in yellow) preceding the respective failure (red flash) is created.

For the root cause analysis method, we could use our presented PU learning approach and train a model with two classes: \emph{positive} class $\mathcal{P}$ with normal data, and \emph{unknown} class $\mathcal{U}$ containing both normal and root cause log lines (yellow in Figure~\ref{fig:problem_example}). This setup would involve numerous inaccurate labels in class $\mathcal{U}$ since most samples are considered normal log lines.

After assigning log lines to $\mathcal{P}$ or $\mathcal{U}$, we could train our machine learning model. It employs our objective function for PU learning to predict \emph{root cause scores} for log lines within $\mathcal{U}$.

\subsection{Balancing Data to Boost Performance on Rare Cases}
\label{sec:balancing}
Real-world system failures can arise from various root causes, resulting in imbalanced training data. Insufficient samples of specific root causes in $\mathcal{U}$ can hinder the model's understanding of their distribution. This bias towards common cases may lead to challenges in distinguishing actual root cause log lines from those in the normal class $\mathcal{P}$.

In an unsupervised context, addressing class imbalance is complex due to the lack of distribution information. 
To mitigate this problem, we could estimate root cause numbers through automatic clustering and subsequently balance the data accordingly.

To balance the training data automatic clustering could be used to estimate root cause numbers and occurrences in the training dataset. Each cluster would then represents a different root cause type. This process enhances training for underrepresented root causes while ensuring a focus on common ones.

\begin{figure}[htbp]
\centering
\includegraphics[width=0.7\columnwidth]{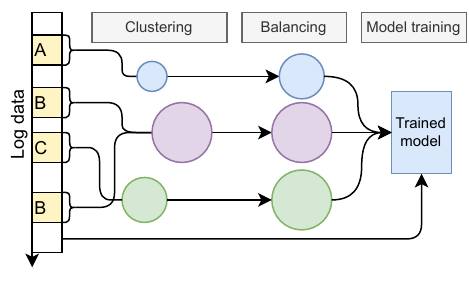}
\caption{Balancing the training data.}
\label{fig:balancing}
\end{figure}

Figure~\ref{fig:balancing} depicts the outlined balancing process. Initially, time windows (yellow on the left) are encoded into a vector $\vec{w}$ using meta-information from log lines within those windows. The information from which service the logs $x_i$ originate, making each vector a representation of the involved services for each root cause. The vector $\vec{w}$'s dimensionality equals the different values of $x_i$:
$$dim(\vec{w}) = |\{\forall x_i \in \mathcal{L}: unique(x_i)\}|$$

Each cluster in the automatic clustering output would represent a specific root cause. While this step is not perfectly accurate, it should provide a reasonable estimate to balance the training data. For instance, in our example, 'B' occurs twice, and 'A' and 'C' once each. The circle's size indicates the log lines within the corresponding time windows of a cluster. This way, log lines from 'B's time windows are combined.

In the balancing step, log lines in each cluster of clustering $\mathcal{K}$ are counted: $U = {|k_i|:\forall k_i \in \mathcal{K}}$. These counts are then normalized between $\frac{max(U)}{2}$ and $max(U)$. In this process, the smallest cluster would have half as many log lines as the largest. 
This ensures that rare root causes are not underrepresented but common root causes can still be learned more efficiently.
The target size $t(\cdot)$ of each cluster $k_i$ could be computed using the following equation:

\begin{equation}
t_{|k_i|} = \frac{|k_i| - min(U)}{max(U) - min(U)} \cdot (max(U) - \frac{max(U)}{2}) + \frac{max(U)}{2})
\end{equation}

In the equation, $|k_i|$ represents the log line count in each cluster. We aim to normalize it within the desired range of $\frac{max(U)}{2}$ and $max(U)$. The smallest cluster corresponds to $min(U)$, and the largest to $max(U)$.

Balancing the training data is crucial. It enables a root cause analysis model to thoroughly train for various root causes while ensuring proper training for the most common ones, which are generally more frequent. As a result, class $\mathcal{U}$ is now balanced, while class $\mathcal{P}$ remains untouched.

\subsection{Envisioning Root Cause Analysis in Complex IT Systems}
In the realm of root cause analysis, PU learning offers a visionary approach that holds the potential to uncover underlying triggers of system failures. Since complex IT systems are continuously monitored, and anomalies are automatically detected, we should not stop at the point of detection. The system should seamlessly transition into a mode of proactive reaction.

With PU learning, the need for laborious and accurate labeling of every log message is diminished. The system intelligently learns from a mixture of positive (normal) and unlabeled (potentially anomalous) data, adapting to the inherent uncertainties and variations in real-world IT environments. As anomalies are detected, the system not only alerts the operations team but also initiates an automated process of identifying potential root causes.

This vision entails an AI-driven ecosystem where the system autonomously learns patterns and relationships across vast amounts of log data. It dynamically creates hypotheses about potential root causes based on the detected anomalies and their contextual information. The system can rapidly hypothesize multiple scenarios, evaluating them against historical root cause patterns that are in the clusters, thus narrowing down the possibilities to the most probable root causes.

These root causes can be verified by system experts and therefore feed into a training loop.
This iterative process continually refines the system's learning process, enabling it to adapt and learn from new root causes over time.

In conclusion, the integration of PU learning into root cause analysis paints a picture of a proactive, automated, and intelligent approach. It empowers IT teams to swiftly understand and mitigate system failures, translating into enhanced reliability, reduced downtime, and optimize operational performance in the complex landscape of modern IT systems.

\section{Conclusion}\label{sec:6}
In conclusion, this study has shed light on the challenges and advancements in anomaly detection, emphasizing the significance of labeled data, especially for deep learning approaches. The proposed framework has systematically classified log anomalies and explored an automated data labeling strategy to address the labeling dilemma. Additionally, the investigation has extended to diverse anomaly detection techniques, showcasing their applicability across various anomaly types. Beyond anomaly detection, the study envisions the evolution of root cause analysis as a consequential phase after anomaly identification. This unexplored territory holds the promise of deeper insights into the triggers behind anomalies, with profound implications for effective IT system management.

\bibliographystyle{IEEEtran}
\bibliography{bib}

\end{document}